%% file: acl2018.tex
\documentclass[11pt,a4paper]{article}
\usepackage[hyperref]{acl2018}
\usepackage{times}
\usepackage{amssymb}
\usepackage{amsmath}
\usepackage{amsfonts}
\usepackage{amsthm}
\usepackage{url}
\usepackage{times}
\usepackage{latexsym}
\usepackage{subcaption}
\usepackage[inline]{enumitem}
\usepackage{multirow}
\usepackage{booktabs}
\usepackage[english]{babel}
\usepackage{array}
\usepackage{epsfig}
\usepackage{enumitem}
\usepackage{graphicx}
\usepackage{caption}
\usepackage{subcaption}
\usepackage{algorithm,algpseudocode}

\aclfinalcopy 

\title{Transfer Learning for Context-Aware Question Matching in\\ Information-seeking Conversations in E-commerce}
\author{{\bf Minghui Qiu$^1$, Liu Yang$^2$, Feng Ji$^1$, Weipeng Zhao$^1$, Wei Zhou$^1$}\\
  {\bf Jun Huang$^1$, Haiqing Chen$^1$, W. Bruce Croft$^2$, Wei Lin$^1$ }\\
  $^1$Alibaba Group, Hangzhou, China \\
  $^2$Center for Intelligent Information Retrieval, University of Massachusetts Amherst \\
  {\tt \{minghui.qmh,zhongxiu.jf\}@alibaba-inc.com}\\
  {\tt \{lyang,croft\}@cs.umass.edu}\\
}

\date{}

\begin{document}
\maketitle
\begin{abstract}
Building multi-turn information-seeking conversation systems is an important and challenging research topic. Although several advanced neural text matching models have been proposed for this task, they are generally not efficient for industrial applications. Furthermore, they rely on a large amount of labeled data, which may not be available in real-world applications.
To alleviate these problems, we study transfer learning for multi-turn information seeking conversations in this paper. We first propose an efficient and effective multi-turn conversation model based on convolutional neural networks. After that, we extend our model to adapt the knowledge learned from a resource-rich domain to enhance the performance. Finally, we deployed our model in an industrial chatbot called AliMe Assist~\footnote{
Interested readers can access AliMe Assist through the Taobao App, or the web version via \url{https://consumerservice.taobao.com/online-help}}
and observed a significant improvement over the existing online model.
\end{abstract}

\section{Introduction}
\label{sec:intro}


With the popularity of online shopping, there is an increasing number of customers seeking information regarding their concerned items. To efficiently handle  customer questions, a common approach is to build a conversational customer service system~\cite{alime-demo,liu-sigir}. 
In the E-commerce environment, the information-seeking conversation system can serve millions of customer questions per day. According to the statistics from a real e-commerce website~\cite{qiu:acl17}, the majority of customer questions (nearly 90\%) are business-related or seeking information about logistics, coupons etc. 
Among these conversation sessions, $75\%$ of them are more than one turn\footnote{According to a statistic in AliMe Assist in Alibaba Group}.  
Hence it is important to handle multi-turn conversations or context information in these conversation systems. 

Recent researches in this area have focused on deep learning and reinforcement learning \cite{DBLP:conf/acl/ShangLL15,DBLP:conf/sigir/YanSW16,DBLP:conf/acl/LiGBSGD16,DBLP:conf/emnlp/LiMRJGG16,DBLP:conf/naacl/SordoniGABJMNGD15,DBLP:conf/acl/WuWXZL17}. 
One of these methods is Sequential Matching Network\cite{DBLP:conf/acl/WuWXZL17}, which matches a response with each utterance in the context at multiple levels of  granularity and leads to state-of-the-art performance on two multi-turn conversation corpora. However, such methods suffer from at least two problems: they may not be efficient enough for industrial applications, and they rely on a large amount of labeled data which may not be available in reality.


To address the problem of efficiency, we made three major modifications to SMN to boost the efficiency of the model while preserving its effectiveness. First, we remove the RNN layers of inputs from the model; Second, SMN uses a Sentence Interaction based (SI-based) Pyramid model~\cite{pang:aaai2016} to model each utterance and response pair. In practice, a Sentence Encoding based (SE-based) model like BCNN~\cite{yin:naacl2015} is complementary to the SI-based model. Therefore, we extend the component to incorporate an SE-based BCNN model, resulting in a hybrid CNN (hCNN)~\cite{jianfei-wsdm18}; Third, instead of using a RNN to model the output representations, we consider a CNN model followed by a fully-connected layer to further boost the efficiency of our model. As shown in our experiments, our final model yields comparable results but with higher efficiency than SMN.

To address the second problem of insufficient labeled data, we study transfer learning (TL)~\cite{pan:tkde2010} to utilize a source domain with adequate labeling to help the target domain. 
A typical TL approach is to use a shared NN~\cite{mou:EMNLP2016,yang:iclr2017} and domain-specific NNs to derive shared and domain-specific features respectively. 
Recent studies~\cite{ganin:jmlr2016,taigman:iclr2017,ChenSQH17,liu:acl2017} consider adversarial networks to learn more robust shared features across domains.
Inspired by these studies, we extended our method with a Transfer Learning module to leverage information from a resource-rich domain. Similarly, our TL module consists of a shared NN and two domain-specific NNs for source and target domains. The output of the shared NN is further linked to an adversarial network as used in~\cite{liu:acl2017} to help learn domain invariant features. Meanwhile, we also use domain discriminators on both source and target features derived by domain-specific NNs to help learn domain-specific features. Experiments show that our TL method can further improve the model performance on a target domain with limited data.

To the best of our knowledge, our work is the first to study transfer learning for context-aware question matching in conversations. Experiments on both benchmark and commercial data sets show that our proposed model outperforms several baselines including the state-of-the-art SMN model. We have also deployed our model in an industrial bot called AliMe Assist~\footnote{\url{https://consumerservice.taobao.com/online-help}} and observed a significant improvement over the existing online model.

\input{model}

\input{exp}

\section{Related Work}
Recent research in multi-turn conversations has focused on deep learning and reinforcement learning \cite{DBLP:conf/acl/ShangLL15,DBLP:conf/sigir/YanSW16,DBLP:conf/acl/LiGBSGD16,DBLP:conf/emnlp/LiMRJGG16,DBLP:conf/naacl/SordoniGABJMNGD15,DBLP:conf/acl/WuWXZL17,liu-sigir}. 
The recent proposed Sequential Matching Network (SMN) \cite{DBLP:conf/acl/WuWXZL17} matches a response with each utterance in the context at multiple levels of granularity, leading to state-of-the-art performance on two multi-turn conversation corpora. Different from SMN, our model is built on CNN based modules, which has comparable results but with better efficiency. 

We study transfer learning (TL)~\cite{pan:tkde2010} to help domains with limited data. TL has been extensively studied in the last decade. With the popularity of deep learning, many Neural Network (NN) based methods are proposed~\cite{yosinski:nips2014}. A typical framework uses a shared NN to learn shared features for both source and target domains~\cite{mou:EMNLP2016,yang:iclr2017}. Another approach is to use both a shared NN and domain-specific NNs to derive shared and domain-specific features~\cite{liu:acl2017}. 
This is improved by some studies~\cite{ganin:jmlr2016,taigman:iclr2017,ChenSQH17,liu:acl2017} that consider adversarial networks to learn more robust shared features across domains.
Our TL model is based on~\cite{liu:acl2017}, with enhanced source and target specific domain discrimination losses.

\section{Conclusion}
\label{sec:conclusion}
In this paper, 
we proposed a conversation model based on Multi-Turn hybrid CNN (MT-hCNN). We extended our model to adapt knowledge learned from a resource-rich domain.
Extensive experiments and an online deployment in AliMe E-commerce chatbot showed the efficiency, effectiveness, and transferablity of our proposed model.

\section*{Acknowledgments}
The authors would like to thank Juwei Ren, Zhiyu Min and other members of AliMe team for their help in deploying our model online. This work was supported in part by NSF grant IIS-1419693. We would also like to thank reviewers for their valuable comments.

\bibliography{acl2018}
\bibliographystyle{acl_natbib}

\end{document}

%% file: model.tex
\section{Model}\label{sec:model}

Our model is designed to address the following general problem. 
Given an input sequence of utterances $\{u_1, u_2, \ldots, u_n\}$ and a candidate question $r$, our task is to identify the matching degree between the utterances and the question. When the number of utterances is one, our problem is identical to paraphrase identification (PI)~\cite{yin:naacl2015} or  natural language inference (NLI)~\cite{snli:emnlp2015}. Furthermore, we consider a transfer learning setting to transfer knowledge from a source domain to help a target domain. 



\subsection{Multi-Turn hCNN (MT-hCNN)}
\label{basic-model}

\begin{figure}[th!]
\centering
\includegraphics[width=0.7\columnwidth]{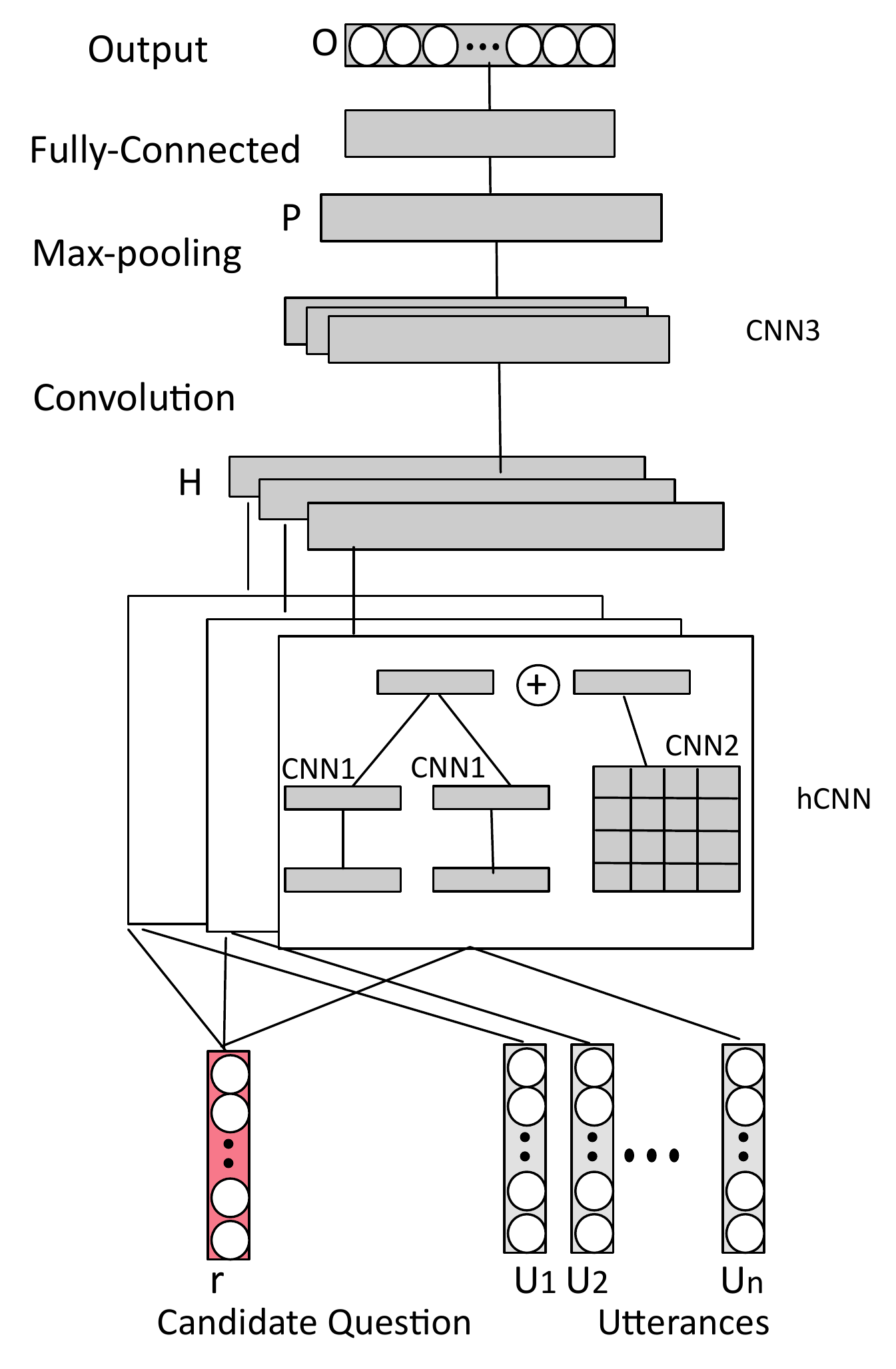}
\caption{Our proposed multi-turn hybrid CNN.}
\label{fig:model}
\end{figure}

We present an overview of our model in Fig.~\ref{fig:model}. In a nutshell, our model first obtains a representation for each utterance and candidate question pair using hybrid CNN (hCNN), then concatenates all the representations, and feeds them into a CNN and fully-connected layer to obtain our final output.

The hybrid CNN (hCNN) model~\cite{jianfei-wsdm18} is based on two models: a modified SE-based BCNN model~\cite{yin:tacl2016} and a SI-based Pyramid model~\cite{pang:aaai2016}. The former encode the two input sentences separately with a CNN 
and then combines the resulting sentence embeddings as follows:
\begin{eqnarray*}
\mathbf{h_1} &=& \text{CNN}_1(\mathbf{X}_{1}); \quad \mathbf{h_2} = \text{CNN}_1(\mathbf{X}_{2}). \\
\mathbf{H_b} &=& \mathbf{h_1} \oplus \mathbf{h_2} \oplus (\mathbf{h_1}-\mathbf{h_2}) \oplus (\mathbf{h_1} \cdot \mathbf{h_2}).
\end{eqnarray*}
where `$-$' and `$\cdot$' refer to element-wise subtraction and multiplication, and `$\oplus$' refers to concatenation.

Furthermore, we add a SI-base Pyramid component to the model, we first produce an interaction matrix $\mathbf{M} \in \mathbb{R}^ {m \times m}$, where $\mathbf{M_{i,j}}$ denotes the dot-product score between the $i^\text{th}$ word in $\mathbf{X}_1$ and the $j^\text{th}$ word in $\mathbf{X}_2$.
Next, we stack two 2-D convolutional layers and two 2-D max-pooling layers on it to obtain the hidden representation $\mathbf{H_p}$. Finally, we concatenate the hidden representations as output for each input sentence pair:$\quad \mathbf{Z_{X_1,X_2}} = \text{hCNN}(X_1, X_2) = \mathbf{H_b} \oplus \mathbf{H_p}$.

We now extend hCNN to handle multi-turn conversations, resulting MT-hCNN model. Let $\{u_1,u_2,u_3,\ldots,u_n\}$ be the utterances, $r$ is the candidate question.
\begin{eqnarray*}
\mathbf{h_{u_i,r}} &=& \text{hCNN}(\mathbf{u_i}, r).   \quad  for   \quad  i \in [1,n] \\
H &=& [h_{u_1,r}; h_{u_2,r}; \cdots ;h_{u_n,r}].\\
P &=& \text{CNN}_3(H).\\
O &=& \text{Fully-Connected}(P)
\end{eqnarray*}
Note that $H$ is obtained by stacking all the $\mathbf{h}$, CNN$_3$ is another CNN with a 2-D convolutional layer and a 2-D max-pooling layer, the output of CNN$_3$ is feed into a fully-connected layer to obtain the final representation $O$. 

\subsection{Transfer with Domain Discriminators}
We further study  transfer  learning (TL) to learn knowledge from a source-rich domain to help our target domain, in order to reduce the dependency on a large scale labeled training data.
As similar to~\cite{liu:acl2017}, we use a shared MT-hCNN and two domain-specific MT-hCNNs to derive shared features $\mathbf{O^c}$ and domain-specific features $\mathbf{O^s}$ and $\mathbf{O^t}$. The domain specific output layers are:
\begin{eqnarray}
 \hat{y}^k =
\begin{cases}
\sigma (\mathbf{W^{sc}} \mathbf{O^c}+ \mathbf{W^s} \mathbf{O^s} + \mathbf{b^s}), \; \text{if $k=s$}\\
\sigma (\mathbf{W^{tc}} \mathbf{O^c}+ \mathbf{W^t} \mathbf{O^t} +\mathbf{b^t}), \; \text{if $k=t$}\\
\end{cases}
\end{eqnarray}
where $\mathbf{W^{sc}}$, $\mathbf{W^{tc}}$, $\mathbf{W^{s}}$, and $\mathbf{W^{t}}$ are the weights for shared-source, shared-target, source, and target domains respectively, while $\mathbf{b^s}$ and $\mathbf{b^t}$ are the biases for source and target domains respectively.

%

Following~\cite{liu:acl2017}, we use an adversarial loss $L_{a}$ 
to encourage the shared features learned to be indiscriminate across two domains:
\begin{align}
\label{eqn:advloss}
L_{a} = & \frac{1}{n} \sum^{n}_{i=1}\sum_{d\in {s,t}} p(d_i=d | \mathbf{U},r) \log p(d_i=d | \mathbf{U},r). \notag 
\end{align}
where $d_i$ is the domain label and $p(d_i | \cdot)$ is the domain probability from a domain discriminator.

Differently, to encourage the specific feature space to be discriminable between different domains, we consider applying domain discrimination losses on the two specific feature spaces. We further add two negative cross-entropy losses: $L_s$ for source and $L_t$ for target domain: 
\begin{align}
L_{s} = & -\frac{1}{n_s} \sum^{n_s}_{i=1} \mathbb{I}^{d_{i}=s} \log p(d_{i}=s | \mathbf{U}^s,r^s). \notag\\
L_{t} = & -\frac{1}{n_t} \sum^{n_t}_{i=1} \mathbb{I}^{d_{i}=t} \log p(d_{i}=t | \mathbf{U}^t,r^t). \notag
\end{align}
where $\mathbb{I}^{d_{i}=d}$ is an indicator function set to 1 when the statement ($d_{i}=d$) holds, or 0 otherwise.

Finally, we obtain a combined loss as follows: 
\begin{align}
\notag\mathcal{L} =  & \sum_{\textbf{k} \in {\mathbf{s},\mathbf{t}}} - \frac{1}{n_{\textbf{k}}}\sum^{n_{\textbf{k}}}_{j=1} \frac{1}{2}(y^k_j - \hat{y}^k_j)^2 + 
\frac{\lambda_1}{2} L_{a} \\
\notag & + \frac{\lambda_2}{2} L_{s} + \frac{\lambda_3}{2} L_{t}
 + \frac{\lambda_4}{2} ||\mathbf{\Theta}||^2_F.
\end{align}
where $\mathbf{\Theta}$ denotes model parameters. 

%% file: exp.tex
\begin{table*}[th!]
 	\centering
 	\small
 	\caption{Comparison of base models on Ubuntu Dialog Corpus (UDC) and an E-commerce data (AliMe).}
 	\label{tab:exp_res_udc}
 	\begin{tabular}{l|l l l l l | l l l l l}
 		\hline
 		Data     & \multicolumn{5}{c|}{UDC} & \multicolumn{5}{c}{AliMeData} \\ \hline
Methods	&	MAP	&	R@5	&	R@2	&	R@1	&	Time	&	MAP	&	R@5	&	R@2	&	R@1	&	Time	\\\hline
ARC-I	&	0.2810	&	0.4887	&	0.1840	&	0.0873	&	16	&	0.7314	&	0.6383	&	0.3733	&	0.2171	&	23	\\
ARC-II 	&	0.5451	&	0.8197	&	0.5349	&	0.3498	&	17	&	0.7306	&	0.6595	&	0.3671	&	0.2236	&	24	\\
Pyramid	&	0.6418	&	0.8324	&	0.6298	&	0.4986	&	17	&	0.8389	&	0.7604	&	0.4778	&	0.3114	&	27	\\
Duet	&	0.5692	&	0.8272	&	0.5592	&	0.4756	&	20	&	0.7651	&	0.6870	&	0.4088	&	0.2433	&	30	\\\hline
MV-LSTM	&	0.6918	&	0.8982	&	0.7005	&	0.5457	&	1632	&	0.7734	&	0.7017	&	0.4105	&	0.2480	&	2495	\\
SMN	&	\textbf{0.7327}	&	\textbf{0.9273}	&	0.7523	&	0.5948	&	64	&	0.8145	&	0.7271	&	0.4680	&	0.2881	&	91	\\\hline
MT-hCNN-d	&	0.7027	&	0.8992	&	0.7512	&	0.5838	&	20	&	0.8401	&	0.7712	&	0.4788	&	0.3238	&	31	\\
MT-hCNN	&	0.7323	&	0.9172	&	\textbf{0.7525}	&	\textbf{0.5978}	&	24	&	\textbf{0.8418}	&	\textbf{0.7810}	&	\textbf{0.4796}	&	\textbf{0.3241}	&	36	\\
\hline
 	\end{tabular}
 \end{table*}
 
\section{Experiments}
\label{exp}
We evaluate the {efficiency} and {effectiveness} of our base model, the {transferability} of the model, and the {online evaluation} in an industrial chatbot.

\noindent  \textbf{Datasets:}
We evaluate our methods on two multi-turn conversation corpus, namely Ubuntu Dialog Corpus (UDC)~\cite{DBLP:journals/corr/LowePSP15} and AliMe data.

\textbf{Ubuntu Dialog Corpus:} The Ubuntu Dialog Corpus (UDC) \cite{DBLP:journals/corr/LowePSP15} contains multi-turn technical support conversation data collected from the chat logs of the Freenode Internet Relay Chat (IRC) network. We used the data copy shared by Xu et al. \cite{DBLP:journals/corr/XuLWSW16}, in which numbers, urls and paths are replaced by special placeholders. It is also used in several previous related works \cite{DBLP:conf/acl/WuWXZL17,liu-sigir}\footnote{The data can be downloaded from \url{https://www.dropbox.com/s/2fdn26rj6h9bpvl/ubuntu\%20data.zip?dl=0}}. It consists of $1$ million context-response pairs for training, $0.5$ million pairs for validation and  $0.5$ million pairs for testing.

\textbf{AliMe Data:} We collect the chat logs between customers and a chatbot called AliMe from ``2017-10-01'' to ``2017-10-20'' in Alibaba~\footnote{The textual contents related to user information are filtered.}. 
The chatbot is built based on a question-to-question matching system~\cite{alime-demo}, where for each query, it finds the most similar candidate question in a QA database and return its answer as the reply. It indexes all the questions in our QA database using Lucene\footnote{\url{https://lucene.apache.org/core/}}. For each given query, it uses TF-IDF ranking algorithm to call back candidates. 
To form our data set, we concatenated utterances within three turns~\footnote{Around $85\%$ of conversations are within 3 turns.} to form a query, and used the chatbot system to call back top 15 most similar candidate questions as candidate ``responses''.~\footnote{A ``response'' here is a question in our system.} We then asked a business analyst to annotate the candidate responses, where a ``response'' is labeled as positive if it matches the query, otherwise negative. In all, we have annotated 63,000 context-response pairs. This dataset is used as our \textit{Target} data. 

Furthermore, we build our \textit{Source} data as follows. In the AliMe chatbot, if the confidence score of answering a given user query is low, i.e. the matching score is below a given threshold\footnote{The threshold is determined by a business analyst}, we prompt top three related questions for users to choose. We collected the user click logs as our source data, where we treat the clicked question
as positive and the others as negative. We collected 510,000 query-question pairs from the click logs in total as the source. For the source and target datasets, we use 80\% for training, 10\% for validation, and 10\% for testing. 

\noindent  \textbf{Compared Methods:}
We compared our multi-turn model (MT-hCNN) with two CNN based models ARC-I and ARC-II~\cite{DBLP:conf/nips/HuLLC14}, and several advanced neural matching models: MV-LSTM~\cite{DBLP:conf/aaai/WanLGXPC16}, Pyramid~\cite{pang:aaai2016} 
{Duet}~\cite{Mitra:2017:LMU:3038912.3052579}, SMN~\cite{DBLP:conf/acl/WuWXZL17}\footnote{The results are based on the TensorFlow code from authors, and with no over sampling of negative training data.}, and a degenerated version of our model that removes CNN$_3$ from our MT-hCNN model (MT-hCNN-d).
All the methods in this paper are implemented with TensorFlow and are trained with NVIDIA Tesla K40M GPUs.

\noindent  \textbf{Settings:}
\label{exp-setting}
We use the same parameter settings of hCNN in~\cite{jianfei-wsdm18}. For the $\text{CNN}_3$ in our model, we set window size of convolution layer as 2, ReLU as the activation function, and the stride of max-pooling layer as 2. The hidden node size of the \text{Fully-Connected} layer is set as 128.
AdaDelta is used to train our model with an initial learning rate of 0.08. We use MAP, Recall@5, Recall@2, and Recall@1 as evaluation metrics. We set $\lambda_1=\lambda_2=\lambda_3=0.05$, and $\lambda_4=0.005$.

\subsection{Comparison on Base Models}\label{eval:open}
The comparisons on base models are shown in Table~\ref{tab:exp_res_udc}. First, the RNN based methods like MV-LSTM and SMN have clear advantages over the two CNN-based approaches like ARC-I and ARC-II, and are better or comparable with the state-of-the-art CNN-based models like Pyramid and Duet; Second, our MT-hCNN outperforms MT-hCNN-d, which shows the benefits of adding a convolutional layer to the output representations of all the utterances; 
Third, we find SMN does not perform well in AliMeData compared to UDC. One potential reason is that UDC has significantly larger data size than AliMeData (1000k vs. 51k), which can help to train a complex model like SMN; Last but not least, our proposed MT-hCNN shows the best results in terms of all the metrics in AliMeData, and the best results in terms of R@2 and R@1 in UDC,  which shows the effectiveness of MT-hCNN.

We further evaluate the inference time~\footnote{The time of scoring a query and N candidate questions, where N is 10 in UDC, and 15 in AliMeData.} of these models. 
As shown in Table~\ref{tab:exp_res_udc}, MT-hCNN has comparable or better results when compared with SMN (the state-of-the-art multi-turn conversation model), but is much more efficient than SMN ($\sim$60\% time reduction). 
MT-hCNN also has similar efficiency with CNN-based methods but with better performance. As a result, our MT-hCNN module is able to support a peak QPS~\footnote{Queries Per Second} of 40 on a cluster of 2 service instances, where each instance reserves 2 cores and 4G memory on an Intel Xeon E5-2430 machine. This shows the model is applicable to industrial bots. In all, our proposed MT-hCNN is shown to be both efficient and effective for question matching in multi-turn conversations. 

\subsection{Transferablity of our model}\label{eval:tl}
To evaluate the effectiveness of our transfer learning setting, we compare our full model with three baselines: Src-only that uses only source data, Tgt-only that uses only target data, and TL-S that uses both source and target data with the adversarial training as in~\cite{liu:acl2017}. 

As in Table~\ref{tab:exp3}, Src-only performs worse than Tgt-only. This shows the source and target domains are related but different. 
Despite the domain shift, TL-S is able to leverage knowledge from the source domain and boost performance; Last, our model shows better performance than TL-S, this shows the helpfulness of adding domain discriminators on both source and target domains.
\begin{table}[h!]
\centering
\caption{Transferablity of our model.}
\label{tab:exp3}
\begin{tabular}{l|l l l l}
	\hline
	Data & \multicolumn{4}{c}{E-commerce data (AliMeData)} \\ \hline
	Methods  & MAP    & R@5 & R@2  & R@1 \\ \hline 
Src-only	&	0.7012	&	0.7123	&	0.4343	&	0.2846	\\
Tgt-only	&	0.8418	&	0.7810	&	0.4796	&	0.3241	\\
TL-S 	&	0.8521	&	0.8022	&	0.4812	&	0.3255	\\
Ours	&	\textbf{0.8523}	&	\textbf{0.8125}	&	\textbf{0.4881}	&	\textbf{0.3291}	\\
\hline
	\end{tabular}
\end{table}

\subsection{Online Evaluations}\label{eval:online}
We deployed our model online in AliMe Assist Bot. For each query, the bot uses the TF-IDF model in Lucene to return a set of candidates, then uses our model to rerank all the candidates and returns the top. We set the candidate size as 15 and context length as 3. To accelerate the computation, we bundle the 15 candidates into a mini-batch to feed into our model. We compare our method with the online model - a degenerated version of our model that only uses the current query to retrieve candidate, i.e. context length is 1. We have run 3-day A/B testing on the Click-Through-Rate (CTR) of the models. 
As shown in Table~\ref{tab:online}, our method consistently outperforms the online model, yielding $5\% \sim 10\%$ improvement.
\begin{table}[h!]
\centering
\caption{Comparison with the online model.}
\label{tab:online}
\begin{tabular}{l|l l l }
	\hline
CTR  & Day1    & Day2 & Day3 \\ \hline
Online Model	&	0.214	&	0.194   & 0.221 \\
Our Model       &	\textbf{0.266}   &   \textbf{0.291}   & \textbf{0.288} \\
\hline
\end{tabular}
\end{table}
\vspace{-0.1in}